\documentclass[11pt,a4paper]{article}
\usepackage[utf8]{inputenc}
\usepackage{geometry}
\usepackage{natbib}
\usepackage{url}

\title{AI in data science education: experiences from the classroom}
\author{J.A. Hageman* and C.F.W. Peeters \\
\\
Mathematical and Statistical Methods Group (Biometris),\\
Wageningen University \\
P.O. Box 16, 6700 AA, Wageningen, The Netherlands \\
\\
*Corresponding author: jos.hageman@wur.nl}
\date{}

\begin{document}

\maketitle

\begin{abstract}
This study explores the integration of AI, particularly large language models (LLMs) like ChatGPT, into educational settings, focusing on the implications for teaching and learning. Through interviews with course coordinators from data science courses at Wageningen University, this research identifies both the benefits and challenges associated with AI in the classroom. While AI tools can streamline tasks and enhance learning, concerns arise regarding students' overreliance on these technologies, potentially hindering the development of essential cognitive and problem solving skills. The study highlights the importance of responsible AI usage, ethical considerations, and the need for adapting assessment methods to ensure educational outcomes are met. With careful integration, AI can be a valuable asset in education, provided it is used to complement rather than replace fundamental learning processes.
\end{abstract}


\section{Introduction}
ChatGPT, developed by OpenAI, is a large language model (LLM) that uses deep learning to understand and generate human-like text based on the input it receives \citep{openai2023}. LLMs are artificial intelligence (AI) systems trained on vast amounts of text data to understand and generate human-like language. LLMs are a subset of and a key application within AI. LLMs are capable of performing a wide range of tasks, from answering questions and summarizing information to generating creative content. 

AI / LLMs have also found their way into educational settings, starting basically a new era in education \citep{shen2024,parker2024,gill2024}. A change that can be compared to e.g., the shift from relying on paper-based abstract books to online databases for literature queries. Since the introduction of ChatGPT at the end of 2022, students worldwide have now access to a plethora of AI tools that can assist students with diverse tasks. These tools can assist with research \citep{cavazos2024}, provide explanations \citep{cavazos2024,gill2024}, and even help with creative writing and coding tasks \citep{shen2024}, all within seconds. 

The use of these tools in an educational setting presents some challenges. One concern is students not meeting the intended learning outcomes for courses they participate in. The fear is that overreliance on AI could hinder the development of cognitive and problem-solving skills in students \citep{vargas2023}. Note that in the past the introduction in education of new technologies or new practices has evoked similar concerns \citep{ellis2023}. In that light, we could compare the pitfalls from the introduction of AI in education to the introduction of the calculator \citep{ellis2023}, computers and computerized exercises \citep{ringstaff2002}, solution manuals \citep{georgieva2002}, flipped classroom \citep{akcayir2018}, group work, and group assignments \citep{benson2019}. These innovations initially faced (or still face) similar resistance but have eventually become much encountered practices in education. 

There are also other concerns, for example the risk of plagiarism, as students might rely on AI-generated content for assignments \citep{cavazos2024}. Another problem is that AI models can generate wrong or biased results \citep{cavazos2024}, something that students might not be able to recognize (yet) \citep{ellis2023}. Also, the use of AI could widen educational inequalities, as these tools may not be available for all students, putting some students at a disadvantage \citep{gill2024}. Lastly, the environmental impact by AI is deemed problematic \citep{george2023}.

The initial response from educators was cautionary at best \citep{ellis2023}. Their responses ranged from warning students against possible plagiarism and not meeting learning outcomes for their courses, to downright banning any use of AI in course work \citep{gill2024}.

LLMs have the unique possibility to generate text and responses autonomously and this does introduce challenges, as outlined above, in educational settings \citep{parker2024}. At the same time, new technologies hardly disappear after their introduction because of initial objections. This underscores the need for educators to carefully integrate LLMs into their courses. The goal is to facilitate students meeting the intended learning outcomes and speed up the development of cognitive and problem-solving skills in students, allowing educators to spend more time on relevant topics or even adding new topics to their courses. This approach could be the educational equivalent of the Olympic motto: “faster, higher, stronger – together”.

To effectively integrate AI into educational settings, it can be helpful to assess and understand current practices and experiences \citep{cavazos2024,vargas2023}. By learning from existing approaches, educators can build on successful strategies and avoid common pitfalls. It is useful to talk to educators who have begun noticing or even started incorporating AI tools into their courses. Through personal interviews, this paper has collected the views on AI in the classroom from a set of course coordinators within the data science education landscape at Wageningen University, The Netherlands. The goal was to analyze how the emergence of AI tools like ChatGPT is influencing their courses, particularly in relation to learning outcomes, teaching practices, and assessment methods. Through these insights, this paper aims to identify common trends, challenges, and opportunities in integrating AI into the educational process.

\section{Methodology}

\subsection{Wageningen University}
Wageningen University is a public university in the Netherlands which research and education revolves around 5 themes: Climate Change, Biodiversity, Feeding the World, Circular Economy and Healthy Food \& Living. Artificial intelligence runs like a thread through all the themes \citep{wur2024}. To support this, Wageningen University offers an increasing choice of education possibilities in data science through a range of different data science courses and data science tracks embedded in different bachelor and master programs.

\subsection{Interviews}
Ten data science courses were selected from a total of 34 available data science courses at Wageningen University. These 10 courses were selected in such a way that they spanned a wide range of study programs and served a varied number of students per course per year (student numbers per course per year: min: 7, mean: 104, max: 504). The course coordinators from these 10 data science courses were invited by email to an approximately 1-hour interview. In the invitation some example questions related to the influence of AI on their courses were given. Of those 10 course coordinators, 8 interviews were ultimately conducted. One course coordinator declined and an another interview was not pursued further due to unavailability of the interviewee. Of the completed interviews, 1 interview was conducted via a questionnaire, 1 interview was held online and the remaining ones were held in person.
	
The interviews were structured in the sense that there were predefined topics to be discussed (for instance observed changes in the classroom, changes in student skills, changes in student behavior, use of AI in the course, implemented changes in the course with respect to e.g., examination, ethics, etc.). For each topic starting questions were prepared, some of which were already shared in the invitation email. During the interviews there was also the possibility for exploration, depending on the answers given by the interviewees. Using these follow-up questions, interviewees were encouraged to share (more of) their personal insights and where possible, to give specific examples. During the interviews notes were taken which were transcribed immediately after the interview. 

\subsection{Analysis}
Once the interviews were completed, a systematic analysis was applied to the collected data to retrieve meaningful insights. All transcripts were carefully read, noting significant points, key themes, similarities and other noteworthy insights raised by the course coordinators. The analysis focused on uncovering key themes that spanned across multiple interviews. These themes ranged from practical uses of AI in the classroom to broader ethical concerns, as well as the role of instructors in guiding AI use. In addition to identifying similarities, differences in perspectives were also captured. These findings served as the foundation for the next section of this paper, which discusses the commonalities, differences, and other significant findings from all the interviews.

\section{Results}
The following section presents insights and trends gathered from the interviews with the course coordinators. This section highlights the consensus on the impact of AI on their courses and learning outcomes. This section emphasizes statements and ideas that reflect common experiences and practices shared among the interviewed group, focusing on identifying widespread similarities and trends across different courses.

\subsection{AI as a tool}
AI tools are broadly recognized as a valuable educational tool by the educators across the various courses. Especially for courses that require students to write scripts, as LLMs can assist students in understanding and improving their code. For instance, students can use LLMs to generate code snippets for their programming tasks or they can copy-paste their error messages into AI to help debug existing code. In some courses, teachers have hypothesized, students are helped by LLMs in improving the quality of the written work by enhancing their vocabulary and grammar. For some teachers it feels like they spent less time on improving the English of written texts, although other initiatives like thesis rings also contribute to this. This group of educators recognizes that LLMs can also act as a study coach for students, by continuously asking questions and zooming in on the specific parts of course material students find difficult, students’ understanding can greatly be improved. This is something a single educator with many students and a limited number of contact hours cannot provide. The overarching theme is that LLMs can significantly streamline and support the learning process by speeding up routine tasks and providing instant feedback.

Despite its advantages, there is a consensus in the group of educators that AI should not be used with the core learning activities as specified in the learning goals. The primary goal in a course is for students to engage with the material and develop essential skills independently as specified in the learning outcomes. Over-reliance on AI for completing exercises or (graded) assignments will hinder the development of problem-solving abilities. Ideally, students should use AI at a minimum in activities that are related to the primary learning goals of courses, forcing students to think for themselves. Second best is students scarcely using AI but making sure they understand all AI outputs before continuing. Both are difficult to control, and this group of educators does not like to take on a policing role. 

\subsection{Responsible use and ethical considerations of AI}
The group of educators emphasizes the importance of students being aware of how to use AI responsibly. They encourage thoughtful and ethical use of AI tools, which includes understanding when to use AI and when to avoid it. For instance, AI should not be used for primary learning outcomes, such as writing code for modeling purposes if that is the course’s main goal. However, using AI for tasks not closely connected to the core teaching objectives, like visualizing results, is acceptable to them if it does not undermine the primary learning outcomes. Students should critically evaluate and verify AI-generated results instead of blindly trusting them. This is particularly crucial for primary learning outcomes where students are first encountering the topic. This emphasizes the importance of students not using AI for the main goals of their courses as students likely lack the necessary experience to judge the correctness of AI output in these cases.

Transparency in AI usage is another key aspect. The group of educators would like that students disclose their use of AI in their assignments and projects, including a number of concrete examples of how they utilized these tools. This transparency helps maintain academic integrity and allows instructors to accurately assess the students' understanding. 

Ethical concerns surrounding AI use are a significant focus. These concerns include the ownership of content generated by AI and the responsibility for errors in AI outputs. The group of educators feel that students need to understand that while AI can assist in producing work, the ultimate responsibility for the work lies with them. They must learn to identify and correct errors in AI-generated outputs to ensure they are not overly dependent on AI and can independently verify the accuracy of their work. Class wide demonstrations of the errors AI currently makes in assignments can be of great value as it shows to students how subtle these can be and the confidence with which AI can produce wrong results. Additionally, the use of AI raises concerns about plagiarism and the authenticity of student work. 

By making students aware of how undermining the use of AI can be in reaching primary learning goals and also of ethical concerns, this group of educators hope that students develop the correct attitude towards AI use, so they use it responsibly without the need for educators to spend a lot of time policing students.

\subsection{Assessment adaptations}
To ensure the authenticity of students' work and their understanding of the material, several adaptations to assessment methods are being implemented. Oral exams are very useful for verifying students' knowledge levels. These exams allow instructors to probe the depth of students' understanding and assess their ability to e.g. explain concepts without the aid of AI. This direct interaction helps educators judge students’ skills in a very direct manner, without students relying on AI-generated content which could happen with take-home assignments.

Another significant adaptation is the introduction of on-the-spot assignments. These tasks require students to complete their work in a controlled environment, ensuring that the submitted work is genuinely their own. This method is very effective in preventing the use of AI during take home assignments. The assignments that are created in a controlled environment, with only approved tools, help educators to better judge if students attained the (primary) learning goals in a course.

A third option is to create AI proof assignments. Educators can devise questions in such a way that AI is not likely to provide the correct answers when students use AI for a particular assignment. Although possible and used now, this feels like an arms race that will probably be won by AI.

In principle, the requirement for students to document their use of AI in their assignments (including examples) should also allow educators to judge if primary learning goals were met by the student. However, if students do not have the responsibility to achieve primary learning goals by themselves, it remains to be seen if they will be transparent about this in their AI disclosure.

\subsection{AI for educators}
AI is also there for the benefit of educators. It can improve various aspects of teaching and learning by students. It supports student learning by helping them understand complex concepts, generate code or text, and provide additional explanations, which helps overcome learning barriers and improves the educational experience. 

Instructors use AI to create teaching materials such as lecture slides, summaries, and problem sets for exercises or exam questions. This reduces workload for educators and at the same time not sacrificing quality. AI-generated resources can easily be updated and customized, allowing for dynamic and responsive teaching strategies.

\section{Discussion}
The influence of AI on education is significant and growing. AI offers numerous benefits in supporting and enhancing the learning process, from LLMs assisting with coding and writing to creating teaching materials and grading assignments. Focus for educators is on the following two aspects. (1) Students need to develop essential (core) skills before using AI as a support tool. For example, students must first learn to code themselves before using AI to optimize their code. (2) Students might use AI for refining and correcting their work, but they need to maintain complete control over the process and outcomes. Students should be able to critically analyze AI outputs and make informed decisions based on their understanding. 

While AI can supplement (or replace) lower-order cognitive skills (such as remembering and understanding in Bloom’s taxonomy \citep{bloom1956}), it is crucial that students achieve these skills first to be able to progress to higher-order skills (such as analyzing, evaluating, and creating). Ensuring that students do not skip essential foundational learning steps is a significant concern.

Establishing clear guidelines for ethical AI is important. This includes addressing concerns related to the ownership of AI-generated work and responsibility for errors. This can be done on a course basis or in a wider setting.

By emphasizing responsible use and adapting assessment methods, educators can ensure that AI serves as a powerful tool for learning without compromising the development of essential skills and ethical standards. Education in the future will most likely involve AI, and the role of AI will likely increase. It is crucial that both students and educators are prepared for this new situation.

\section*{Data Availability Statement}
The participants of this study did not give written consent for their data to be shared publicly, so due to the sensitive nature of the research supporting data is not available.

\section*{Conflict of interest}
No conflict of interests.

\bibliographystyle{apalike}
\bibliography{references}

\end{document}